\documentclass{article} 
\usepackage{iclr2024_conference,times}

\usepackage{amsmath,amsfonts,bm}

\def\eqref#1{equation~\ref{#1}}

\def\1{\bm{1}}

\DeclareMathAlphabet{\mathsfit}{\encodingdefault}{\sfdefault}{m}{sl}
\SetMathAlphabet{\mathsfit}{bold}{\encodingdefault}{\sfdefault}{bx}{n}

\usepackage{hyperref}
\usepackage{url}

\usepackage{graphicx}
\usepackage{amsmath}
\usepackage{amssymb}
\usepackage{mathtools}
\usepackage{amsthm}
\usepackage{bm}

\title{Hijacking Context in Large Multi-modal\\ Models}

\author{Joonhyun Jeong\textsuperscript{\rm 1,\rm 2} \\
NAVER Cloud, ImageVision\textsuperscript{\rm 1}\\ Korea Advanced Institute of Science and Technology (KAIST)\textsuperscript{\rm 2} \\
\texttt{\{joonhyun.jeong\}@navercorp.com}
}

\iclrfinalcopy 
\begin{document}

\maketitle

\begin{abstract}
    Recently, Large Multi-modal Models (LMMs) have demonstrated their ability to understand the visual contents of images given the instructions regarding the images. Built upon the Large Language Models (LLMs), LMMs also inherit their abilities and characteristics such as in-context learning where a coherent sequence of images and texts are given as the input prompt. However, we identify a new limitation of off-the-shelf LMMs where a small fraction of incoherent images or text descriptions mislead LMMs to only generate biased output about the hijacked context, not the originally intended context. To address this, we investigate whether replacing the hijacked visual and textual contexts with the correlated ones via GPT-4V and text-to-image models can help yield coherent responses. 
\end{abstract}
\section{Introduction}

 Large Language Models (LLMs) pre-trained with huge amount of text corpus have demonstrated remarkable generalization on various language tasks such as human-like dialogues~\cite{google_2023, openai2023, openai2023gpt4}, reasoning-tasks~\cite{wei2021finetuned, lewkowycz2022solving, yao2022react}, and few-shot in-context learning~\cite{min2021metaicl, min2022rethinking} where a few demonstration examples are given as a context for answering the test question. By bootstrapping LLMs with several add-on layers, Large Multi-modal Models (LMMs)~\cite{openai2023gpt4, fromage, llava, minigpt} were enabled to understand the visual contents and answer the corresponding textual instruction. LMMs also inherited the in-context learning capability of LLM, where a coherent sequence of images and textual information are given as a context for the input prompt~\cite{fromage, openai2023gpt4}.

\begin{figure*}[t]
    \centering
    \includegraphics[width=1.1\linewidth]{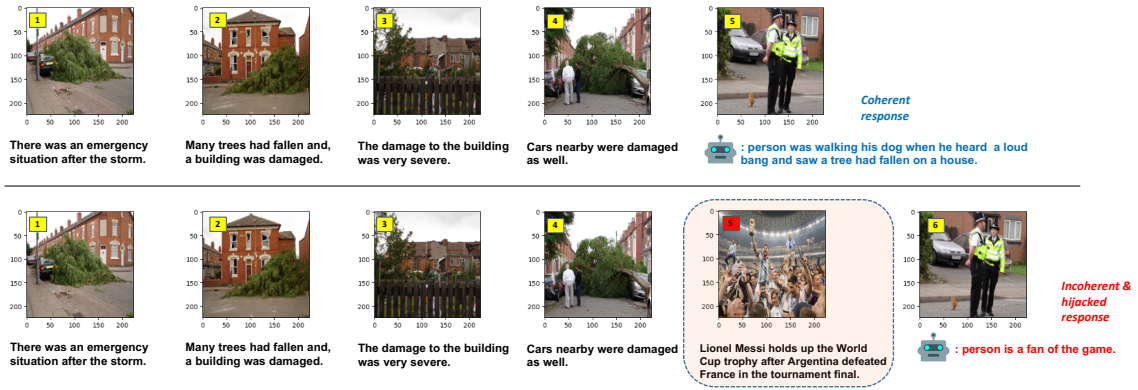}

    \vspace{3mm}
    \caption{\textbf{Hijacking context confuses LMMs to only generate responses with incoherent contents.} (top): Given a sequence of visual and textual story plot 1$\sim$4, LMM reasonably outputs a coherent caption for the final image. (bottom): When a single pair of incoherent image and caption (highlighted with red) is appended to the context, LMM only tells about the hijacked context ($i.e.,$ \textit{football game}), disregarding all the aforementioned context of visual story plots ($i.e.,$ \textit{emergency situation}). 
    }
    \vspace{3mm}
    \label{fig:teaser}
\end{figure*}

 In this paper, we identify a new limitation of these off-the-shelf LMMs, \textit{hijacking contexts}, where a small fraction of incoherent images or text descriptions mislead them to only generate biased output about the incoherent contexts. In Figure \ref{fig:teaser}, we provide the qualitative example of context hijacking on visual story telling (VIST) dataset~\cite{vist} with FROMAGe~\cite{fromage} as LMM. Given all the former sequence of images and their corresponding captions, the LMM is asked to provide a caption or description of the final query image. Normally, when all the visual and textual information is coherent and consistently aligned, LMM successfully provides a coherent response under consideration in the given contexts. However, when only a single image-text pair that contains an irrelevant subject is appended to the context, LMM falls into the hijacked context and loses coherency concerning the bunch of formerly given original contexts. In real-world applications and scenarios of LMMs, there is no guarantee for the inexistence of such noises and irrelevant contexts. For reliable usage of LMM, it should be robust to such distribution shift and stick to the majority of context without confusion on the minority of irrelevant contexts. 
 

To address this, we investigate whether modifying the hijacked image and text contexts with the correlated ones can help yield better coherent responses. Firstly, we instructed GPT-4 to identify any irrelevant sentences and replace them with appropriate alternative sentences that convey coherency given in the contexts. Subsequently, we prompted text-to-image diffusion models~\footnote{We employed DALLE-3~\cite{dalle3} for its detailed sensitiveness to the input prompts} to generate a coherent image corresponding to the newly replaced sentence under consideration for the original sequence of sentences and images.
As a result, a new alternative image and caption that are coherent with the original story plot are interleaved into inputs for LMMs. 
Notably, we observed that LMMs still exhibit hijacked responses, due to two possible reasons: 1) the generated captions are still confusing which seem to be the combination and mixture between the original and hijacked context. 2) the generated images are not realistic and do not have the consistent texture or styles with the original image sequences.

We showcase our problem statement and aforementioned methods and investigations with qualitative examples one by one\footnote{We tested with FROMAGe~\cite{fromage} as the baseline of LMM.}. In summary, our key contributions are outlined as follows:
\begin{itemize}
    \item{Identifying a new problem of off-the-shelf LMMs, context hijacking, that can deteriorate reliable usage of LMMs.}
    \item{Propose a straightforward remedy for suppressing hijacked contexts via the enormous foundation models.}
    \item{Investigate an open question of whether replacing hijacked contexts can help to produce coherent responses, promoting a new future research frontier.}
\end{itemize}



\section{Related Works}

\paragraph{Large language models.} Recently, large language models (LLMs) have gained considerable versatility based on enlarging the scale of pre-training datasets~\cite{gpt-2, gpt-3, openai2023gpt4}. Along with the development of human-aligned feedbacks~\cite{rlhf}, LLMs have further been well-suited for conforming to the instruction rather than predicting simple meaningless next tokens and words. These LLMs have also demonstrated versatility on various language domain tasks including in-context learning~\cite{gpt-3, chan2022data} and long-term dialogues~\cite{jozefowicz2016exploring, dai2019transformer, liu2023lost}. 

\paragraph{Large multi-modal models.} Despite the compelling performance of LLMs on language tasks, LLMs could not understand the visual contexts in an image. To bootstrap LLMs with vision modality, several works~\cite{llava, minigpt, fromage} employed an additional pre-trained visual encoder and leveraged a few linear layers to adapt the visual tokens to be understandable by the frozen LLMs. FROMAGe~\cite{fromage} further demonstrated the in-context learning capability of LMMs where consecutive sequences of image-text pairs are given as a context to respond to the query image. In this paper, we focus on the vulnerability of LMMs to incoherent and irrelevant context information. 

\paragraph{Hijacking context in language and vision tasks.} In the language domain, several works empirically observed that LLMs are easily hijacked by irrelevant context information~\cite{shi2023large}, small perturbations of characters ($e.g.,$ from "film" to "fi1m")~\cite{wang2023adversarial}, and adversarial addition of suffix words~\cite{qiang2023hijacking}. In the vision domain, there have been several approaches that investigated the adversarial robustness of LMMs~\cite{schlarmann2023adversarial, qi2023visual} where a small adversarial tweak of pixel values significantly misled LMMs to output violent content. 
To the best of our knowledge, our work is the first to investigate hijacking contexts in an in-context learning scenario where the consecutive sequence of images and texts along with irrelevant ones are given as the context to provide answers about the query image.

\section{Context-Hijacking in Large Multi-modal Models}

\paragraph{Preliminary: Large Language Models.}
An autoregressive LLM $p_{\theta}$ is pre-trained on a large corpus of text tokens with the next-token prediction objective. Formally, given a text of image caption $y$ and its sequence of tokens ($\bm{s} = s_1, ..., s_{t-1}$) separated by a BPE tokenizer~\cite{sennrich2015neural}, the model is trained to maximize log-likelihood for the next token $s_t$:
\begin{align*}
\log p_{\theta}(y) = \sum_{t=1}^t \log p_{\theta}(s_{t} | \bm{s})
\end{align*}

\paragraph{Preliminary: Large Multi-modal Models.}
To bootstrap LLMs with vision modality, visual embeddings $V_{\phi}(x)$ from input image $x$ are firstly extracted by a visual encoder model $V_{\phi}$ pre-trained on vision tasks. These visual embeddings are mapped into the same dimension of text embedding space by applying several linear layers $W$. These projected visual embeddings, $V_{\phi}(x)^{T}W$, are prepended to the text tokens $s$ and trained to be understandable by the frozen LLMs, letting LLM interpret the image and answer the corresponding captions or instructions. Formally, the image-captioning objective given an image $x$ and a caption $y$ is defined as follows:
\begin{align*}
l(x, y) = \sum_{t=1}^t \log p_{\theta}(s_{t} | V_{\phi}(x)^{T}W, \bm{s})
\end{align*}
For in-context learning scenarios where multiple images and corresponding text descriptions are given, FROMAGe~\cite{fromage} is trained with a random concatenation strategy where multiple image-text pairs are piled for attending and responding more explicitly to each image. Hence, FROMAGe exhibited superior capability for answering the query image $x_n$ given former visual contexts ($x_1, \ldots, x_{n-1}$) and textual contexts $(\bm{s}_1, \ldots, \bm{s}_{n-1})$:
\begin{align*}
\sum_{n=1}^n \log p_{\theta}(s_{n} | V_{\phi}(x_{1})^{T}W, \bm{s}_{1}, \ldots, V_{\phi}(x_{n-1})^{T}W, \bm{s}_{n-1})
\end{align*}

However, when the contexts in the former images ($x_1, \ldots, x_{n-1}$) or texts ($\bm{s}_1, \ldots, \bm{s}_{n-1}$) are composing an incoherent visual story due to the existence of some irrelevant contexts $(x^{*}, \bm{s}^{*})$, the output response tends to be biased towards the irrelevant context, as shown in Figure \ref{fig:teaser}.

\section{Discussions}

\begin{figure}[t]
    \centering
    \includegraphics[width=1.05\linewidth]{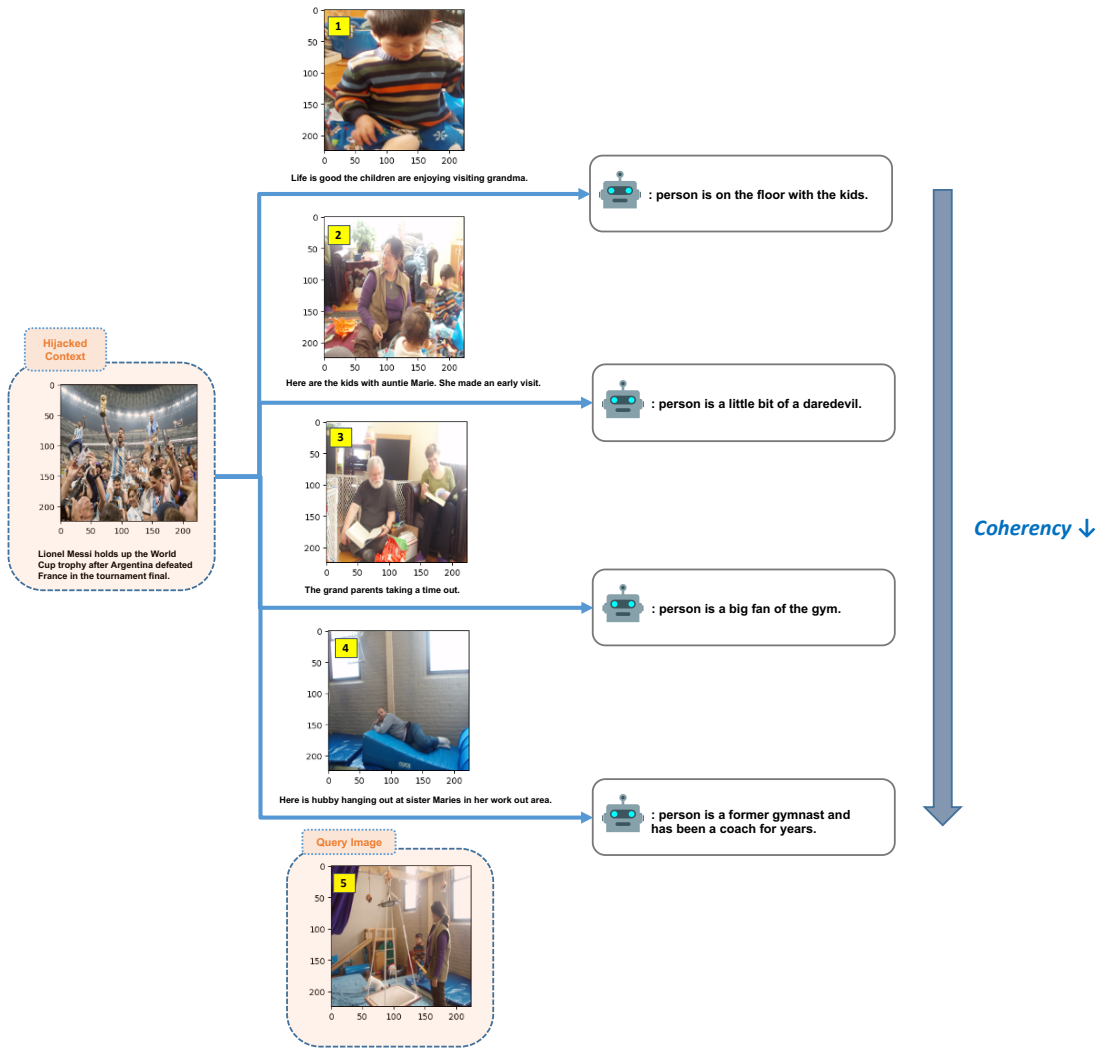}

    \vspace{5mm}
    \caption{\textbf{Effect of location for the hijacking context.} We ablated to insert the hijacking context in between the sequences 1$\sim$5 and visualize the LMM's response with regard to the query image.
    }
    \vspace{2mm}
    \label{fig:ablation_location}
\end{figure}

\paragraph{Location of hijacked context.} We investigate the effect of the sequential location of hijacked context on the coherency of the response from LMMs. In Figure \ref{fig:ablation_location}, when the hijacking context is interleaved in the former part, the final response still tends to adhere to the original context, "\textit{family}". 
In contrast, as the hijacking context is interleaved closer to the query image, the response tends to be incoherent to the former contexts and biased towards the hijacked context, "\textit{football}".

\paragraph{Robustness of GPT-4V.} We further tested GPT-4V whether it is robust to such context-hijacking issues. In Figure \ref{fig:ablation_location}, we inserted hijacked context between sequences 4 and 5 and instructed GPT-4V to answer the caption corresponding to the query image. Notably, GPT-4V's response was not biased towards the hijacked context, rather sticking to the original context:

\textit{Playtime continues with Auntie Marie, as she supervises the little adventurers on their indoor swing set, making memories to cherish.}

This result is indicative of the robustness of GPT-4V toward the distribution shift, shedding light on a future direction of explicitly distilling the robust knowledge from GPT-4V.

\begin{figure}[t]
    \centering
    \includegraphics[width=1.05\linewidth]{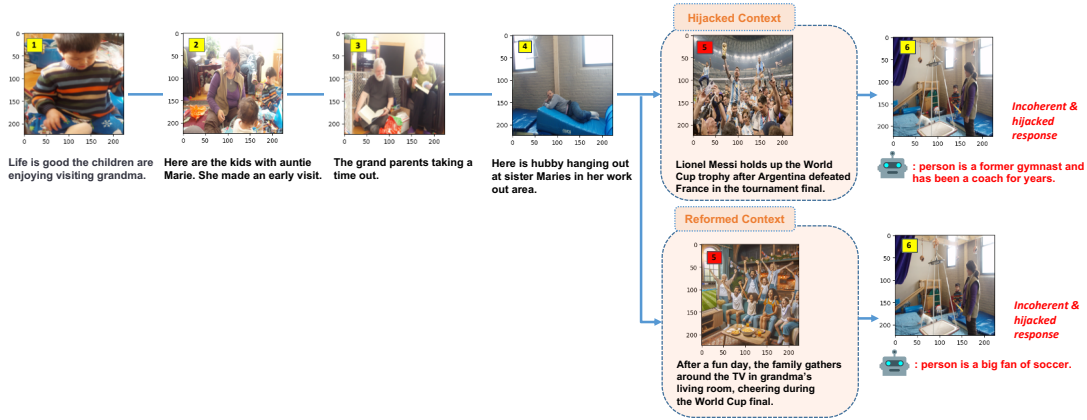}

    \vspace{3mm}
    \caption{\textbf{Effect of reformed context on the LMM response.} We replaced the hijacked context with the one reformed by GPT-4 and DALLE-3.
    }
    \vspace{3mm}
    \label{fig:reformed_response}
\end{figure}

\paragraph{Reforming contexts via large foundation models.} Inspired by robustenss of GPT-4V, we investigate whether such large foundation models can identify and reform the irrelevant visual and textual contexts into coherent ones, for better coherent response. As in Figure \ref{fig:method_reform}, we first instructed GPT-4 to provide appropriate sentences that convey coherency given in the contexts. Subsequently, we instructed DALLE-3 to generate a synthetic image that corresponds to the newly reformed text, with the consideration of underlying visual and textual context information.
\paragraph{Limited impact of reforming hijacked contexts.} In Figure \ref{fig:reformed_response}, we observed that our reforming strategy still does not ensure LMMs give coherent responses. We conjecture that the reformed captions still contain the hijacked context ("\textit{football}") while also adhering to the original context ("\textit{family}"). The generated image also correspondingly contains both visual contexts of "family" and "football", confusing LMMs to produce biased responses. We conjecture that the style and texture of the reformed image differ from the original images, which might also induce LMMs to be confused.

\section{Limitations}

The limitations of this work are that we provide only qualitative example results on the VIST dataset since quantitative evaluation is too labor-demanding; the generated texts from LMMs require a human to check if they are aligned with the image and coherent with the story plot. Therefore, we leave thorough and rigorous quantitative evaluation using various LMMs as future work. Also, our proposed scheme is a straightforward solution yet fragile when the irrelevant contexts become the majority within the entire context where even GPT-4 would also be confused. Therefore, developing a more secure way of filtering or replacing a bunch of the hijacked contexts is a promising future research direction. 
\section{Conclusion}
 In conclusion, we newly identified the context-hijacking problem of existing LMM where the irrelevant visual and textual contexts induce the LMM to generate a biased and incoherent response, disregarding the original context. To mitigate this issue, we introduced a simple method to preemptively remove such hijacked contexts and further reform the hijacked contexts into coherent ones, which promotes future research direction to generate more coherent visual and textual information.

\bibliography{iclr2024_conference}
\bibliographystyle{iclr2024_conference}

\newpage
\appendix
\section{Appendix}
\begin{figure*}[h]
    \centering
    \includegraphics[width=1.05\linewidth]{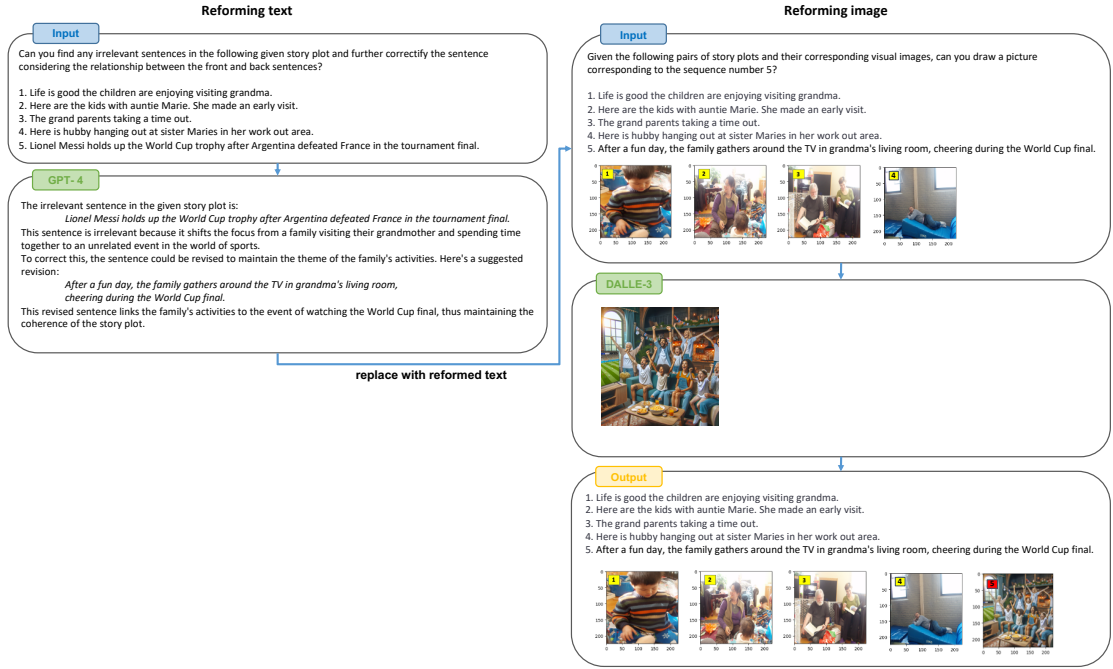}

    \vspace{4mm}
    \caption{\textbf{Reforming hijacked text and images via GPT-4 and DALLE-3.} 
    Given multiple image-text pairs of a visual story plot from VIST dataset~\cite{vist}, GPT-4 is instructed to replace any irrelevant images or text descriptions with coherent ones. Subsequently, DALLE-3 is instructed to generate an image corresponding to the newly reformed text, under consideration of the underlying context.
    }
    \vspace{3mm}
    \label{fig:method_reform}
\end{figure*}

\end{document}